\documentclass[10pt,twocolumn,letterpaper]{article}

\usepackage{cvpr}              
\usepackage{tcolorbox}
\usepackage[table]{xcolor}
\usepackage[accsupp]{axessibility}

\tcbuselibrary{skins,breakable}
\definecolor{lightgold}{RGB}{255, 252, 230}
\definecolor{moonveil}{RGB}{242, 244, 252}
\definecolor{cvprblue}{rgb}{0.21,0.49,0.74}
\usepackage[pagebackref,breaklinks,colorlinks,allcolors=cvprblue,hypertexnames=false]{hyperref}
\newcommand{\tablestyle}[2]{\setlength{\tabcolsep}{#1}\renewcommand{\arraystretch}{#2}\centering\footnotesize}
\def\paperID{} 
\def\confName{CVPR}
\def\confYear{2026}

\title{ViLoMem: Agentic Learner with Grow-and-Refine Multimodal Semantic Memory}

\author{
Weihao Bo$^{1,2}$\quad
Shan Zhang$^{3}$\quad
Yanpeng Sun$^{4}$\quad
Jingjing Wu$^{2}$\quad
Qunyi Xie$^{2}$\quad
Xiao Tan$^{2}$\\
Kunbin Chen$^{2}$\quad
Wei He$^{2}$\quad
Xiaofan Li$^{2}$\quad
Na Zhao$^{4}$\quad
Jingdong Wang$^{2}$ \thanks{Project Leader} \quad
Zechao Li$^{1}$\thanks{Corresponding Author.}\\[.5em]
$^{1}$Nanjing University of Science and Technology\quad
$^{2}$Baidu Inc\\ $^{3}$AIML, Adelaide University\quad $^{4}$Singapore University of Technology and Design \\
{\small \textit{Project page}: \url{https://weihao-bo.github.io/ViLoMeo-page/}}
}

\begin{document}
\maketitle
\begin{abstract}

MLLMs exhibit strong reasoning on isolated queries, yet they operate \emph{de novo}—solving each problem independently and often repeating the same mistakes. 
Current memory agents mainly reuse past trajectories. However, this approach suffers from brevity bias and gradually loses essential domain knowledge. Most critically, even in multimodal tasks, it records only a \textbf{single-modality} trace, failing to capture how visual attention and logical reasoning jointly produce the solution.
This is fundamentally misaligned with human cognition: semantic memory is both \textbf{multimodal and integrated}, preserving visual and abstract knowledge through coordinated but distinct representational streams. We thus introduce \textbf{ViLoMem}, a dual-stream memory framework that constructs compact, schema-based memory. It separately encodes visual distraction patterns and logical reasoning errors, enabling MLLMs to learn from their successful and failed experiences. Following a grow-and-refine principle, the system incrementally accumulates and updates multimodal semantic knowledge—preserving stable, generalizable strategies while avoiding catastrophic forgetting. 
Across six multimodal benchmarks, \textbf{ViLoMem} consistently improves pass@1 accuracy and reduces repeated visual and logical errors. 
Ablations confirm that dual-stream memory is necessary and scalable, supporting long-term learning with efficiency ensured by our retrieval design; learning from visually intensive experiences leads to more faithful multimodal reasoning over time. 




\end{abstract}

\vspace{-1em}    
\section{Introduction}
\label{sec:intro}

Multimodal Large Language Models (MLLMs)~\cite{bai2025qwen2,wu2024deepseek,zhang2025abstractive,zhao2025efficient} have achieved impressive progress in scene understanding~\cite{li2025fvar,wang2025faithfusion,sun2024descriptive,li2024drivingdiffusion}, visual question answering~\cite{jiang2024delving,jiang2024global}, and complex scientific problem solving . Yet despite their growing capability, current MLLMs approach each problem \emph{de novo}—solving every query in isolation, repeatedly re-deriving the same insights and re-committing familiar errors\cite{fang2025comprehensive,zhang2025survey,gao2025survey,tan2025prospect}. Although recent memory-augmented models attempt to mitigate this by storing past interactions~\citep{suzgun2025dynamic, zhang2025agentic}, these memories capture only high-level logical summaries while discarding the visual grounding and perceptual cues essential for multimodal reasoning.

\begin{figure}[t]
    \centering
    \includegraphics[width=0.9\linewidth]{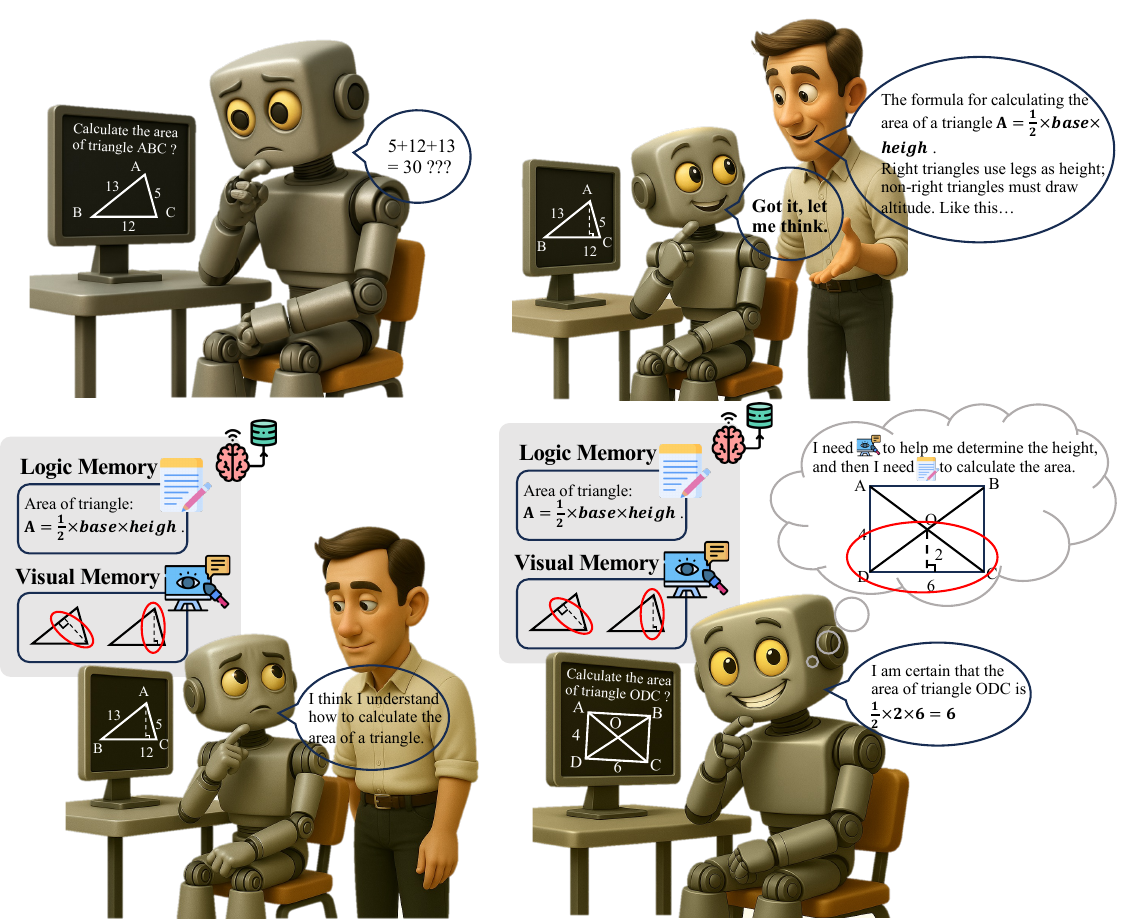}
    \caption{Multimodal Semantic Memory Enables Progressive Learning. When solving multimodal problems, early attempts may contain both logical and visual errors; through feedback, the model refines its logical memory for question-appropriate theorem application and its visual memory to avoid perceptual traps—improving by integrating the \emph{where to look} with the \emph{how to reason.}}
    \label{fig:motivation}
    \vspace{-0.5cm}
  \end{figure}

Recent research has demonstrated that MLLMs' visual perception ability remains fundamentally weaker than their linguistic reasoning, with low-level perceptual failures identified as a primary bottleneck for high-level multimodal reasoning tasks~\citep{lumathvista,sun2025mathglance,tong2024eyes,zhang2025hierarchical}. In mathematical multimodal problem-solving in particular, diagram-perception errors exceed logical reasoning errors, and visual mistakes frequently persist in intermediate reasoning steps even when the final answer is correct~\citep{zhang2025primitive,zhang2025hierarchical}.

This indicates visual attention errors directly cause
downstream logical hallucinations that creates a cascading failure pattern ~\citep{wu2025combating,zhou2025perception}. Our ablation studies further confirm this phenomenon: across six multimodal problem-solving benchmarks, the proportion of visual error summaries consistently exceeds that of logical memory errors (Fig.~\ref{fig:memory_analysis}). Therefore, when solving problems paired with images, it is
essential for models to maintain accurate visual attention to task-relevant regions, avoiding perceptual distractions that propagate into flawed logical inferences.

Logic-only memory is insufficient for multimodal problem solving. While logical theorems and rules are general (e.g., applying the base–height formula for area computation), effective reasoning also requires aligning these abstract rules with their correct visual counterparts (e.g., the shape of triangles). As illustrated in Fig.~\ref{fig:motivation}, triangles exhibit diverse visual configurations, and early attempts may contain both logical and visual errors. Through feedback, the model refines its logical memory for question-appropriate theorem application and its visual memory to avoid perceptual traps, attending to task-relevant regions.
This progressive learning mirrors the human cognitive system, where semantic memory maintains \textit{multimodal representations} that integrate visual experience with abstract reasoning~\citep{dirani2024meg}.

We thus introduce \textbf{ViLoMem}, a dual-stream memory framework that separately models visual distraction patterns and logical hallucination errors as structured schemas, coordinating them through unified retrieval.
Following a grow-and-refine principle, \textbf{ViLoMem} avoids the detail erosion caused by iterative rewriting by filtering similar error patterns and using tailored add/skip and retrieval strategies to incrementally accumulate multimodal semantic knowledge. Specifically, we design custom retrieval strategies for visual and logical streams. For the visual stream, direct image-similarity search is insufficient; the key requirement is helping the model identify question-specific ``visually trapped regions''. To achieve question-aware attention, we generate cross-modal attention maps guided by keywords (previously observed visual mistakes), enabling the model to highlight regions associated with known error patterns relevant to the current question. For the logical stream, instead of directly retrieving query semantically similar logics, the model first analyzes the problem to identify its underlying subject and reasoning requirements—supporting precise positioning of the task type and precise selection of the relevant logical schema.

Overall, \textbf{ViLoMem} automatically attributes successes or failures to the visual or logical stream and updates the corresponding schemas without human supervision. It enables progressive mistake reduction and cross-domain knowledge transfer in multimodal tasks. 
\section{Related Work}
\label{sec:related}

\subsection{Context Engineering}

Recent advancements in agent self-improvement have prominently featured \textit{context engineering}, a paradigm that refines model behavior by strategically modifying input prompts rather than altering the model's underlying weights\cite{agarwal2024many,shao2024scaling,chen2024lifelong,wu2024extending}. These methods primarily leverage natural language feedback, enabling a model to analyze its own performance based on execution traces, reasoning steps, or validation signals and then iteratively revise its operational context~\citep{agrawal2025gepa, shinn2023reflexion, yuksekgonul2024textgrad,wang2025astute}. This approach has given rise to several influential frameworks. For instance, ReAct~\citep{yao2022react} pioneered the integration of reasoning and acting within a synergistic loop. Building on this, Reflexion~\citep{shinn2023reflexion} introduced a mechanism for agents to reflect on past failures, using verbal reinforcement to enhance subsequent planning and decision-making. Other works have focused on optimizing the prompts themselves; TextGrad~\citep{yuksekgonul2024textgrad} proposed a novel method to generate gradient-like textual feedback for prompt refinement, while GEPA~\citep{agrawal2025gepa} demonstrated that an evolutionary approach to prompt optimization based on execution traces can achieve performance surpassing that of traditional reinforcement learning in certain scenarios. However, these approaches are limited by their ephemeral nature; the context is constructed for single interactions, preventing long-term knowledge accumulation. Furthermore, they often suffer from a \textit{brevity bias}~\citep{gao2025prompt}, where iterative refinement strips away crucial details, hindering performance on complex, knowledge-intensive tasks.

\subsection{Long-term Memory}

To address the limitations of transient context, a parallel line of research has focused on equipping agents with \textit{long-term memory}, enabling them to learn from experience and retain knowledge persistently\cite{xu2025mem,alizadeh2024llm,santos2025experimental,zhang2025survey,fan2024videoagent}. This vision is rooted in the cognitive science principle that true, deep learning extends beyond formal training and arises from the continuous accumulation of experience~\citep{ cai2025building,wang2025improving,fang2025lightmem,wu2025evolver}. Research in this area explores various architectures for building durable memory systems. For example, Dynamic Cheatsheet~\citep{suzgun2025dynamic} constructs an external memory that explicitly stores successful and unsuccessful strategies from past inferences, allowing the agent to consult its history. Similarly, ACE~\citep{zhang2025agentic} develops an incremental ``context playbook'' through a generate-reflect-curate cycle, which is designed to avoid the simplification and catastrophic forgetting associated with simple iterative rewriting. The mechanisms for populating these memories are also diverse, ranging from learning through early, formative experiences~\citep{zhang2025agent} and reinforcement learning-based exploration~\citep{zhanglandscape} to interactive learning from noisy, real-time human feedback~\citep{yang2025reinforced, ayub2024interactive}. 

However, these frameworks exhibit a critical blind spot: they are overwhelmingly logic-centric, capturing reasoning patterns while neglecting the visual dimension of multimodal tasks. In contrast, the human brain adopts a hub-and-spoke semantic memory architecture. Visual–semantic associations and error patterns are encoded in the inferotemporal and perirhinal cortex (visual spoke), while abstract reasoning rules and logical error patterns are maintained in the temporal–parietal cortex (logic spoke)\cite{lambon2010coherent, clarke2014object, kuhnke2023role}. The anterior temporal lobe (ATL) serves as the central hub that integrates these modality-specific representations into unified conceptual knowledge. Inspired by this architecture, our AI system implements an \textit{error-aware multimodal semantic memory}, where visual and logical error patterns are stored in separate modality-specific modules, integrated through a semantic hub, and monitored by an executive verifier that detects redundant visual–logical information and modulates attention to prevent recurring mistakes in multimodal scientific reasoning tasks.

\vspace{-0.5em}
\section{Method}

\label{sec:method}

\begin{figure*}[!t]
    \centering
    \includegraphics[width=\textwidth]{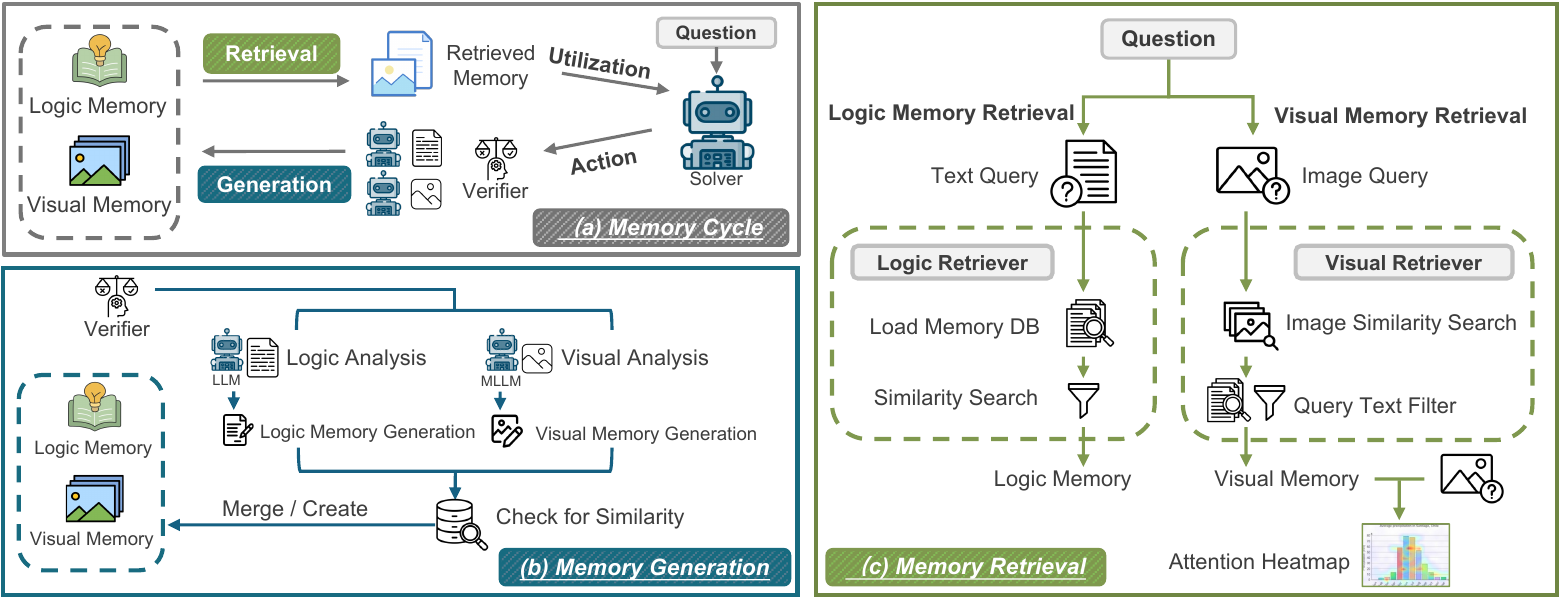}
    \vspace{-1.4em}
    \caption{Overview of the \textbf{ViLoMem} framework. (a)~\textit{Memory Cycle}: A closed-loop learning mechanism where both logical and visual memories are retrieved and utilized by the solver. Retrieval is conditioned on the textual question and its paired image. The solver then performs reasoning steps (actions), which are evaluated by the verifier to filter redundant or invalid trajectories. The remaining trajectories are used to update both memory streams according to their respective types.
    (b)~\textit{Memory Generation}: An error-attribution framework that employs an LLM for logical analysis and an MLLM for visual analysis, producing structured memory schemas through similarity-based \textbf{merge} and \textbf{create} operations. (c)~\textit{Memory Retrieval}: Specialized dual-stream retrieval mechanism. Visual memories undergo a two-stage process involving image-embedding retrieval followed by question-specific retrieval, since visual information must be conditioned on both image content and the textual query. Logical memories are retrieved through problem analysis and text-embedding similarity.}
    \label{fig:method_1}
    \vspace{-0.4cm}
\end{figure*}

We propose \textbf{ViLoMem}, a plug-in dual-stream memory framework for multimodal reasoning in large language models, featuring a closed-loop \textit{Memory Cycle} that enables the agent to continuously learn from its reasoning and perception errors—facilitating progressive, lifelong learning.

\noindent\textbf{Problem Formulation.} Consider a sequence of multimodal inputs $(x_1, x_2, \ldots, x_n)$, where each input $x_i = (I_i, q_i)$ consists of an image $I_i$ and a question text $q_i$. The system maintains two memory banks: a logic memory $\mathcal{M}^L_i = \{m^L_1, m^L_2, \ldots, m^L_{|L|}\}$ storing textual reasoning guidelines, and a visual memory $\mathcal{M}^V_i = \{(m^V_1, I^V_1), (m^V_2, I^V_2), \ldots, (m^V_{|V|}, I^V_{|V|})\}$ storing visual guidelines paired with source images.

As illustrated in Figure~\ref{fig:method_1}(a), the cycle operates as follows: given problem $x_i$, the system performs parallel \textit{Retrieval} from both memory banks to obtain relevant memories $R^L_i$ and $R^V_i$. These retrieved memories are then fed to the \textbf{Solver} for \textit{Utilization}, which generates a candidate answer $\tilde{y}_i$. The \textbf{Verifier} evaluates this answer against the ground truth $y_i$. Upon detecting an error ($\tilde{y}_i \neq y_i$), the system activates the \textit{Generation} process to update both memory banks in parallel, yielding $\mathcal{M}^L_{i+1}$ and $\mathcal{M}^V_{i+1}$. This mechanism enables the agent to progressively refine its capabilities through iterative self-correction.

\noindent\textbf{Core Operations.} We define several key operations used throughout the framework. Let $\phi^T(\cdot)$ and $\phi^M(\cdot)$ denote text and multimodal embedding functions, respectively. The cosine similarity between two embeddings is computed as:
\begin{equation}
    \text{Sim}(u, v) = \frac{u \cdot v}{\|u\| \|v\|}
\end{equation}

For problem analysis during retrieval, we employ an LLM to extract structured information from the question and reasoning trace:
\begin{equation}
    a_i = \text{Analyze}^L(q_i, \tilde{y}_i)
    \label{eq:2}
\end{equation}

The process identifies the problem's subject domain and key concepts. An enriched query is then constructed by combining the original question with this analysis:
\begin{equation}
    \tilde{q}_i = [q_i; a_i]
     \label{eq:3}
\end{equation}

\subsection{Memory Generation}

\label{subsec:memory_generation}

When errors are detected, the system activates a parallel memory-generation framework, as illustrated in Figure~\ref{fig:method_1}(b). This framework conducts detailed error attribution and constructs structured memory units corresponding to two distinct error types.

\subsubsection{Visual Memory Generation}

The visual analysis module, powered by an MLLM, simultaneously identifies the error type and generates corrective guidance. Given the original image $I_i$, question $q_i$, erroneous reasoning trace $\tilde{y}_i$, and ground truth $y_i$, the module produces both error indicator and corresponding guideline within a single model invocation, formally expressed as:
\begin{equation}
    (e^V_i, g^V_i) = \text{AnalyzeGenerate}^V(I_i, q_i, \tilde{y}_i, y_i),
\end{equation}
where $e^V_i \in {\text{True}, \text{False}}$ indicates whether the error originates from visual misinterpretation (e.g., object confusion, overlooked visual symbols, or spatial relationship misunderstandings), and $g^V_i$ denotes the generated \textit{Visual Guideline}—an instruction prescribing the correct observation strategy. All information is stored in a structured JSON dictionary for persistent memory updating. For example, when addressing shape and attribution-related errors in 3D solid objects, the guideline may state: \begin{quote}
``When an object has a uniform, reflective, or metallic-looking surface—even if it appears matte under diffuse lighting—treat it as metallic if it matches the visual style of other known metallic objects in the scene.''
\end{quote}

Before storage, a \textit{similarity check} is performed against existing memories in $\mathcal{M}^V_i$ using text embeddings. The system computes similarity scores $s^V_j = \text{Sim}(\phi^T(g^V_i), \phi^T(m^V_j))$ for all $m^V_j \in \mathcal{M}^V_i$. If $\max_j s^V_j > \tau^V$ (where $\tau^V$ is a similarity threshold), a \textit{merge} operation consolidates the knowledge:
\begin{equation}
    \mathcal{M}^V_{i+1} = \mathcal{M}^V_i \setminus \{(m^V_{j^*}, I^V_{j^*})\} \cup \{(\text{Merge}^V(m^V_{j^*}, g^V_i), I^V_{j^*})\},
\end{equation}
where $j^* = \arg\max_j s^V_j$. Otherwise, a new memory entry is created: $\mathcal{M}^V_{i+1} = \mathcal{M}^V_i \cup \{(g^V_i, I_i)\}$.

\subsubsection{Logical Memory Generation}

In parallel, the logic analysis module, powered by an LLM, examines the reasoning chain for non-visual errors such as computational mistakes, formula misapplications, or logical fallacies. This module focuses solely on textual reasoning without accessing visual information. As formalized in Equation~(6), the module produces both error classification and guideline in a single model invocation:
\begin{equation}
    (e^L_i, g^L_i) = \text{AnalyzeGenerate}^L(q_i, \tilde{y}_i, y_i),
\end{equation}
where $e^L_i \in \{\text{Logical}, \text{Non-Logical}\}$ classifies whether the error involves reasoning failures, and $g^L_i$ represents the abstracted \textit{Logic Guideline}. The model outputs a structured text response containing error type, analysis summary, and guideline fields. For example, when encountering a geometry error arising from incorrect assumptions (i.e., textual biases), the generated guideline may state:
\begin{quote}
``In geometry problems involving perpendicular bisectors, remember that only points lying on the perpendicular bisector segment are guaranteed to be equidistant from the endpoints of the segment. Do not assume a point lies on the bisector unless this is explicitly stated or can be proven from the given construction. Always verify the position of intersection points relative to the bisector before applying the equidistance property.''
\end{quote}

This guideline then undergoes the same \textit{similarity check} and \textit{merge/create} process as visual memory. Similarity scores $s^L_j = \text{Sim}(\phi^T(g^L_i), \phi^T(m^L_j))$ are computed for all $m^L_j \in \mathcal{M}^L_i$, and the memory bank is updated accordingly:
\begin{equation}
    \mathcal{M}^L_{i+1} = \begin{cases}
        \mathcal{M}^L_i \setminus \{m^L_{j^*}\} \cup \{m^L_{\text{new}}\} & s^L_{j^*} > \tau^L \\
        \mathcal{M}^L_i \cup \{g^L_i\} & s^L_{j^*} \leq \tau^L \\
        \mathcal{M}^L_i & \text{otherwise},
    \end{cases}
\end{equation}
where $j^* = \arg\max_j s^L_j$, $m^L_{\text{new}} = \text{Merge}^L(m^L_{j^*}, g^L_i)$, and the update is triggered when $e^L_i = \text{Logical}$ and $g^L_i \neq \emptyset$.

\begin{table*}[tb]
\centering
\caption{Main results across six multimodal reasoning benchmarks. Baseline metrics for Qwen3 series models are sourced from official reports, while GPT-4.1 baselines are from OpenCompass. Metrics marked with * indicate self-evaluated results where official reports are unavailable or show substantial discrepancies. Models with ``(step)'' and ``(+ \textbf{ViLoMem})'' are prompted by step-by-step reasoning.}
\vspace{-.5em}
\label{tab:main_results}
\resizebox{\textwidth}{!}{

\tablestyle{6pt}{1.}
\begin{tabular}{l|cccccc}
\toprule
\textbf{Method} & \textbf{MMMU} & \textbf{MathVista} & \textbf{MathVision} & \textbf{HallusionBench} & \textbf{MMStar} & \textbf{RealWorldQA} \\
\midrule
GPT-4.1 (baseline) & 74.00 & 70.40 & 46.12* & 58.50 & 69.80 & 73.72 \\
GPT-4.1 (step) & 74.16 & 74.27 & 47.47 & 74.44 & 70.43 & 72.03 \\
\rowcolor{lightgold} GPT-4.1 (+ \textbf{ViLoMem}) & \textbf{77.26} & \textbf{76.88} & \textbf{53.95} & \textbf{75.29} & \textbf{72.43} & \textbf{74.38} \\
\midrule
Qwen3-VL-235B-A22B-Instruct (baseline) & 78.70 & 84.90 & 61.28* & 63.20 & 78.40 & 79.30 \\
Qwen3-VL-235B-A22B-Instruct (step) & 75.97 & 83.66 & 62.17 & 74.58 & 76.16 & 78.66 \\
\rowcolor{lightgold} Qwen3-VL-235B-A22B-Instruct (+ \textbf{ViLoMem}) & \textbf{79.40} & \textbf{84.98} & \textbf{62.83} & \textbf{75.21} & \underline{78.31} & 77.22 \\
\midrule
Qwen3-VL-8B-Instruct (baseline) & 66.38* & 77.20 & 48.13* & 61.10 & 70.91 & 71.50 \\
Qwen3-VL-8B-Instruct (step) & 65.52 & 77.80 & 48.35 & 73.08 & 70.22 & 70.85 \\
\rowcolor{lightgold} Qwen3-VL-8B-Instruct (+ \textbf{ViLoMem}) & \textbf{69.90} & \textbf{77.87} & \textbf{49.34} & \textbf{73.19} & \textbf{72.13} & \textbf{73.59} \\
\bottomrule
\end{tabular}
}
\vspace{-1.4em}
\end{table*}

\subsection{Memory Retrieval and Utilization}

\label{subsec:memory_retrieval}

When addressing a new problem $x_i = (I_i, q_i)$, the solver initiates parallel retrieval procedures from both memory banks, as illustrated in Figure~\ref{fig:method_1}(c). Unlike conventional single-stage retrieval, our framework employs specialized strategies for each memory type: visual memory uses a two-stage multimodal-to-text pipeline, while logical memory leverages problem analysis to construct enriched queries.

\subsubsection{Visual Memory Retrieval}

Visual memory retrieval employs a two-stage pipeline that progressively refines candidates from visual similarity to semantic relevance. 

\noindent\textbf{Stage 1: Image Embedding Similarity.} The system first employs multimodal embeddings to compute visual similarity between the query image $I_i$ and all stored memory images. For each memory $(m^V_j, I^V_j) \in \mathcal{M}^V_i$, the similarity is computed as $s^M_j = \text{Sim}(\phi^M(I_i), \phi^M(I^V_j))$. This rapidly recalls a set of top-$k^M$ candidate memories:
\begin{equation}
    \mathcal{C}^V_i = \{(m^V_j, I^V_j) \mid j \in \text{TopK}(\{s^M_j\}, k^M)\}
\end{equation}

\noindent\textbf{Stage 2: Text Embedding Filtering .} Visual similarity alone is insufficient for semantic matching. The system subsequently performs text-based reranking using the enriched query $\tilde{q}_i$ from Equation~(3). For each candidate guideline $m^V_j \in \mathcal{C}^V_i$, text similarity is computed as $s^T_j = \text{Sim}(\phi^T(\tilde{q}_i), \phi^T(m^V_j))$. The final retrieved visual memories are obtained by filtering with threshold $\tau^V$ and selecting top-$k^V$ by similarity score:
\begin{equation}
    R^V_i = \{m^V_j \mid j \in \text{TopK}(\{s^T_j \mid s^T_j \geq \tau^V\}, k^V)\}
\end{equation}

This two-stage process ensures that the retrieved visual memories are both semantically relevant to the current problem and specifically address common visual pitfalls encountered when interpreting similar images.

\noindent\textbf{Focusing on where to look via visual attention maps.} Beyond textual guidelines, we further introduce an auxiliary visual representation of memory cues. Leveraging the retrieved visual memory and its associated error patterns, the system generates question-aware attention maps that highlight historically error-prone regions in the query image $I_i$. These attention maps serve as supplementary visual inputs alongside the original image, providing explicit spatial guidance that directs the model’s focus toward task-relevant areas while avoiding known perceptual traps. Experimental results demonstrate that this visual augmentation yields additional performance improvements (refer to Section~\ref{sec:ablation}).

\subsubsection{Logical Memory Retrieval}
Logical memory retrieval is a text-based semantic matching process. The system constructs an enriched query $\tilde{q}_i$ using Equations~\ref{eq:2}-\ref{eq:3} to capture both the problem text and structured domain information. For each memory $m^L_j \in \mathcal{M}^L_i$, text embedding similarity is computed as $s^L_j = \text{Sim}(\phi^T(\tilde{q}_i), \phi^T(m^L_j))$. The top-$k^L$ most relevant guidelines are retrieved by applying similarity threshold $\tau^L$ and ranking by similarity score:
\begin{equation}
    R^L_i = \{m^L_j \mid j \in \text{TopK}(\{s^L_j \mid s^L_j \geq \tau^L\}, k^L)\}
\end{equation}

\subsubsection{Solution Generation with Dual Memory}
Finally, the solver generates the answer by conditioning on both the original inputs and the retrieved memories from the visual and logical streams:
\begin{equation}
    \tilde{y}_i = \text{Gen}(I_i, q_i, R^L_i, R^V_i),
\end{equation}
where $\text{Gen}$ denotes the MLLM solver that integrates visual perception, question understanding, and dual-stream memory guidance. The retrieved logical guidelines $R^L_i$ provide structured and context-relevant reasoning frameworks, while the visual guidelines $R^V_i$ supply explicit perceptual priors. Together, they enable more robust and accurate multimodal reasoning.
\section{Experiments}
\label{sec:exp}

\subsection{Experimental Setup}

\begin{figure}[tb]
\centering
\includegraphics[width=0.9\columnwidth]{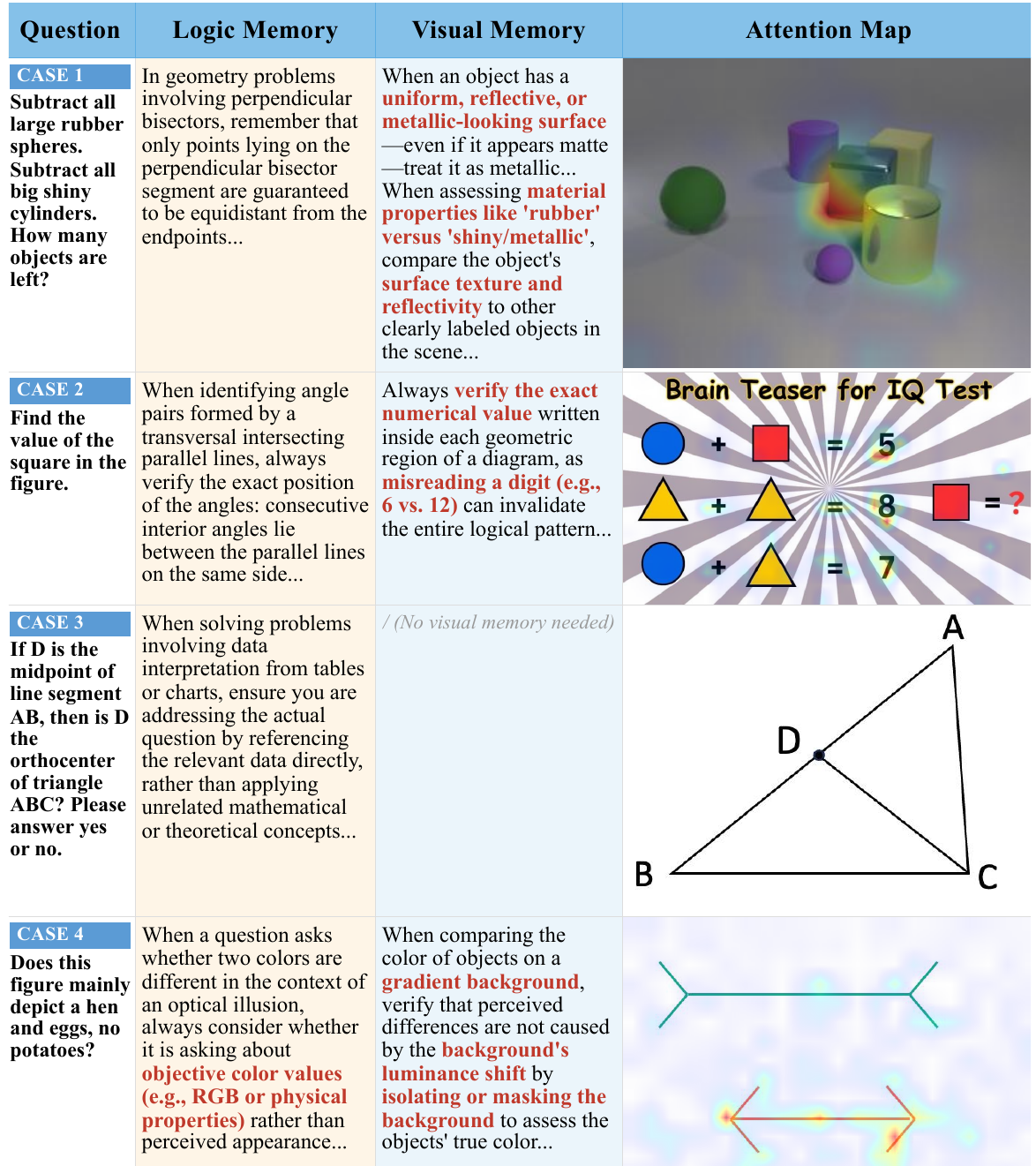}
\vspace{-.4em}
\caption{Visual memory generation and retrieval examples. Each case shows the original error, the extracted visual pattern, and successful retrieval in analogous scenarios.}
\label{fig:visual_memory_showcase}
\vspace{-1.4em}
\end{figure}

\textbf{Tasks and Datasets.}
We evaluate \textbf{ViLoMem} on three multimodal reasoning benchmarks that are particularly sensitive to cumulative visual–logical errors:
(1)~\textit{Hallucination and real-world robustness}, which emphasize language hallucination, visual illusion, and spatial grounding;
(2)~\textit{Multimodal mathematical reasoning}, which couples logic reasoning with visual grounding; and
(3)~\textit{Vision-dependent knowledge}, which requires expert-level visual understanding across multiple disciplines.

{HallusionBench}~\citep{guan2024hallusionbench} diagnoses intertwined language hallucination and visual illusion through 1,129 control-paired questions;
{RealWorldQA}~\citep{xai2024realworldqa} assesses spatial reasoning over 765 natural scenes;
{MathVista (mini)}~\citep{lumathvista} and {MathVision (mini)}~\citep{wang2024measuring} test visual-grounded mathematical reasoning across diverse diagrams and competition-style problems;
{MMMU (val)}~\citep{yue2024mmmu} covers 1050 college-level questions across six academic domains (Art \& Design, Business, Science, Health \& Medicine, Humanities \& Social Science, Tech \& Engineering); and
{MMStar}~\citep{chen2024mmstar} offers 1,500 high-quality samples evaluating vision-dependent reasoning across 18 fine-grained dimensions.

\noindent\textbf{Models and Implementation.} To assess the effectiveness and generalizability of \textbf{ViLoMem}, we evaluate it across models of varying scale and accessibility: the proprietary GPT-4.1 as a strong closed-source baseline, the open-source Qwen3-VL-235B-A22B-Instruct as a state-of-the-art large multimodal model, and Qwen3-VL-8B-Instruct as a smaller model to test whether memory benefits extend to resource-constrained settings. For memory generation, we employ Qwen3-235B-A22B-Instruct for logical memory (pure language reasoning analysis) and Qwen3-VL-235B-A22B-Instruct for visual memory (image-grounded error attribution). Memory retrieval uses Qwen3-Embedding for text similarity and Qwen2.5-VL-Embedding for image similarity, enabling efficient semantic matching. Additional implementation details are provided in the Appendix.

\noindent\textbf{Evaluation Metrics.} We report pass@1 accuracy using VLMEvalKit~\citep{duan2024vlmevalkit}. When rule-based matching detects potential errors, we apply an LLM-as-a-judge mechanism for verification, enhancing scoring accuracy and reducing false negatives from format variations.

\begin{table}[tb]
\centering
\caption{Ablation study on the contribution of dual stream memory components. We evaluate GPT-4.1 with different memory configurations on two representative benchmarks.} 
\label{tab:ablation}
\vspace{-.5em}
\tablestyle{2pt}{1.}
\resizebox{.48\textwidth}{!}{
\begin{tabular*}{\columnwidth}{@{\extracolsep{\fill}} l c c @{}}
\toprule
\textbf{Method} & \textbf{MMMU} & \textbf{MathVista} \\
\midrule
GPT-4.1 (baseline) & 74.00 & 70.40 \\
GPT-4.1 (step) & 74.16 & 74.27 \\
\midrule
GPT-4.1 (w/o logic memory) & 76.64 & 75.59 \\
GPT-4.1 (w/o visual memory) & 76.88 & 75.66 \\
\midrule
\rowcolor{lightgold} GPT-4.1 (+ \textbf{ViLoMem}) & 77.26 & \textbf{76.88} \\
\rowcolor{moonveil} GPT-4.1 (+ \textbf{ViLoMem} \& attention) & \textbf{78.21} & 76.87 \\
\bottomrule
\end{tabular*}}
\vspace{-2.1em}
\end{table}

\subsection{Main Results on Multimodal Benchmarks}

\begin{figure*}[tb]
\centering
\includegraphics[width=\textwidth]{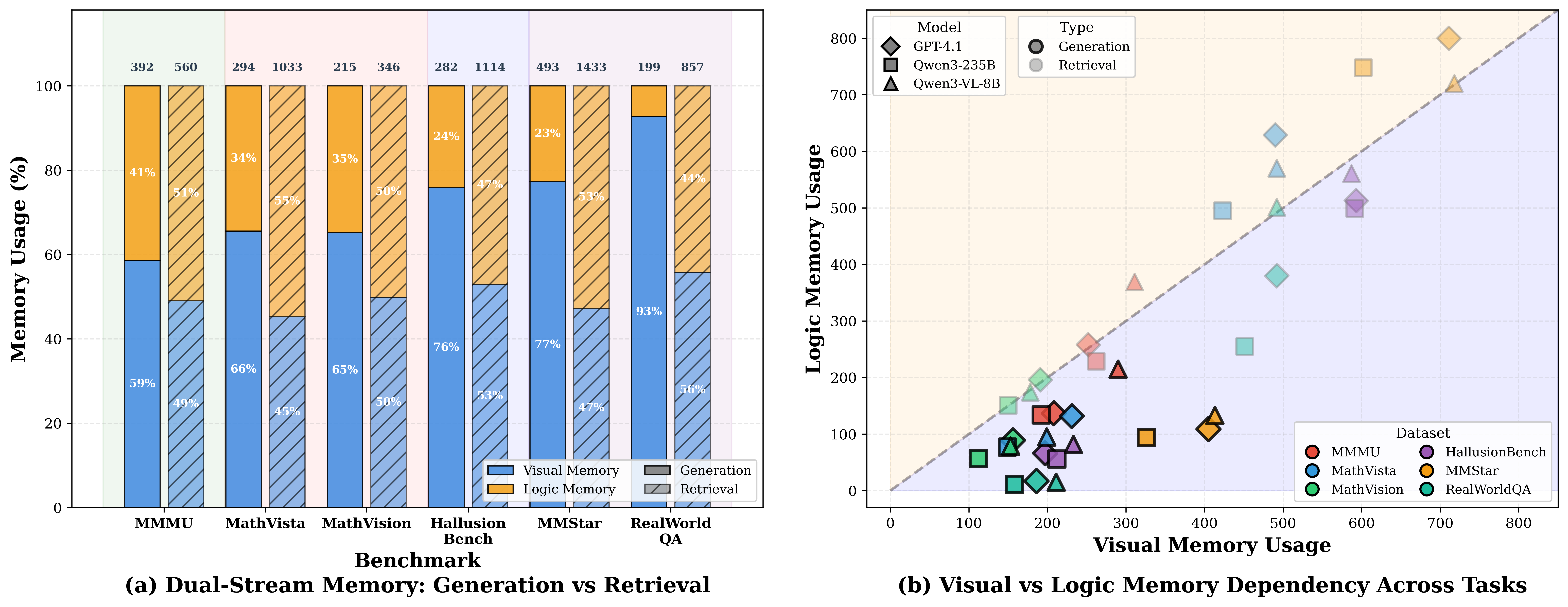}
\vspace{-1.8em}
\caption{Analysis of dual stream memory usage patterns across six benchmarks. (a)~Memory generation and retrieval statistics show that visual errors dominate generation (59\% to 93\%), while retrieval operations significantly exceed generation events. (b)~Cross task dependency analysis reveals balanced utilization of both memory streams during retrieval across diverse tasks and models.}
\label{fig:memory_analysis}
\vspace{-1.2em}
\end{figure*}

Table~\ref{tab:main_results} shows evaluations across six multimodal benchmarks covering mathematical reasoning, hallucination robustness, and visual knowledge understanding. We compare three MLLMs under three configurations:
\textit{Baseline}: following the official default prompting setup;
\textit{Step}: using explicit step-by-step reasoning prompts; and
\textit{+\textbf{ViLoMem}}: integrating our dual-stream memory framework.
The comparison between \textit{Step} and \textit{+\textbf{ViLoMem}} highlights the effectiveness of memory in mitigating \emph{de novo} reasoning and promoting experience-driven problem solving.


\textbf{ViLoMem} achieves consistent improvements across all models, with particularly notable gains on mathematical reasoning benchmarks. This result aligns with our motivation, as mathematical reasoning tasks demand more visually grounded chains of thoughts. Prior studies have shown that visual perception errors significantly degrade reasoning accuracy ~\citep{zhang2025primitive,lumathvista}. By tracking visual errors and integrating them with logical reasoning, \textbf{ViLoMem} effectively enhances overall mathematical reasoning performance.
Among the three MLLMs, {GPT-4.1} shows the largest improvement—particularly on MathVision (+6.48) and MathVista (+2.61)—owing to its stronger contextual learning ability and superior capacity to utilize and interpret past errors for solving similar problems.  Smaller models benefit more substantially from memory augmentation: {Qwen3-VL-8B-Instruct} achieves notable gains on {MMMU} (+4.38) and {RealWorldQA} (+2.74), indicating that structured memory provides complementary knowledge beyond the model’s limited parametric capacity.

Among the evaluated tasks, improvements on knowledge-intensive benchmarks are moderate, as these tasks primarily rely on factual recall rather than multi-step reasoning. Moreover, manual inspection of the stored memory information from both streams reveals two primary performance bottlenecks. First, when the solver exhibits strong textual bias, over-relying on linguistic reasoning while paying limited attention to visual cues, the resulting reasoning traces contain insufficient visual information for the verifier to generate effective visual memory. Second, when the solver struggles to perceive complex diagrams and generates low-quality visual descriptions, the verifier finds it difficult to identify clear visual errors and tends to attribute all errors to the logical stream, often resulting in mixed memory updates. Therefore, a promising direction for future work is to design more specialized mechanisms to further enhance the decoupling the dual memory streams.


\begin{table}[tb]
\centering
\caption{Cross model memory transfer analysis. For each solver, we replace its self-generated memory with memories generated by the other two models on the same benchmark.} 
\vspace{-.5em}
\label{tab:cross_model}
\setlength{\tabcolsep}{5pt}
\tablestyle{8pt}{1.}
\resizebox{.48\textwidth}{!}{
\begin{tabular}{l c c}
\toprule
\textbf{Method} & \textbf{MMMU} & \textbf{MathVista} \\
\midrule
GPT-4.1 {\small(step)} & 74.16 & 74.27 \\
\rowcolor{lightgold} GPT-4.1 {\small(+ \textbf{ViLoMem})} & 77.26 & \textbf{76.88} \\
\rowcolor{moonveil} GPT-4.1 {\small(+ \textbf{ViLoMem} Cross)} & \textbf{78.21} & 76.58 \\
\midrule
Qwen3-VL-235B {\small(step)} & 75.97 & 83.66 \\
\rowcolor{lightgold} Qwen3-VL-235B {\small(+ \textbf{ViLoMem})} & \textbf{79.40} & \textbf{84.98} \\
\rowcolor{moonveil} Qwen3-VL-235B {\small(+ \textbf{ViLoMem} Cross)} & 79.26 & 84.21 \\
\midrule
Qwen3-VL-8B {\small(step)} & 65.52 & 77.80 \\
\rowcolor{lightgold} Qwen3-VL-8B {\small(+ \textbf{ViLoMem})} & 69.90 & 77.87 \\
\rowcolor{moonveil} Qwen3-VL-8B {\small(+ \textbf{ViLoMem} Cross)} & \textbf{71.26} & \textbf{79.20} \\
\bottomrule
\end{tabular}}
\vspace{-1.4em}
\end{table}

\noindent\textbf{Case Study:} Figure~\ref{fig:visual_memory_showcase} illustrates the operation of dual-stream memory in practice. Cases 1, 2, and 4 expose a key limitation of logical memory: it retrieves guidelines irrelevant to the visual context (e.g., recalling perpendicular bisector principles when material discrimination is required). Visual memory effectively addresses this gap by identifying surface reflectivity (Case~1), numerical digits in diagrams (Case~2), and background luminance for color perception (Case~4). The attention maps confirm that retrieved visual cues guide the model toward task-relevant regions (Case~2/~4). Case~3 highlights the plausibility of our memory generation process: when a problem can be solved without visual cues (the question already providing complete visual descriptions), logical memory alone suffices. Overall, visual memory supports perception-intensive tasks, while logical memory governs reasoning-driven problems.

\begin{table*}[htb]
\centering
\caption{Cross benchmark memory generalization analysis. For each benchmark, we exclude its task specific memory and merge memories from all other benchmarks as the retrieval source. Results demonstrate that while cross domain memories provide partial benefits, task aligned memories remain essential for optimal performance.}
\vspace{-.7em}
\label{tab:cross_benchmark}
\tablestyle{9pt}{1.}
\resizebox{\textwidth}{!}{
\begin{tabular}{l c c c c c c}
\toprule
\textbf{Method} & \textbf{MMMU} & \textbf{MathVista} & \textbf{MathVision} & \textbf{HallusionBench} & \textbf{MMStar} & \textbf{RealWorldQA} \\
\midrule
Qwen3-VL-8B (baseline) & 66.38* & 77.20 & 48.13* & 61.10 & 70.91 & 71.50 \\
Qwen3-VL-8B (step) & 65.52 & 77.80 & 48.35 & 73.08 & 70.22 & 70.85 \\
\midrule
\rowcolor{lightgold} Qwen3-VL-8B (+ \textbf{ViLoMem}) & \textbf{69.90} & \textbf{77.87} & 49.34 & \textbf{73.19} & \textbf{72.13} & \textbf{73.59} \\
\rowcolor{moonveil} Qwen3-VL-8B (+ \textbf{ViLoMem} Cross) & 65.14 & 76.10 & \textbf{50.00} & 70.66 & 70.93 & 71.63 \\
\bottomrule
\end{tabular}}
\vspace{-1.8em}
\end{table*}
\subsection{Ablation Study}
\label{sec:ablation}

We validate the necessity of dual-stream memory by selectively disabling each component on GPT-4.1. As shown in Table~\ref{tab:ablation}, removing either stream consistently degrades performance, confirming that both memory types are essential. Removing logical memory leads to larger drops on \textit{MathVista}, where systematic reasoning and formula-related errors frequently recur. In contrast, removing visual memory produces comparable degradation across both benchmarks, indicating that visual distraction errors are pervasive in multimodal reasoning tasks. The gap between the single-stream variants and the full \textbf{ViLoMem} model demonstrates their complementarity: the visual and logical streams capture distinct, rather than redundant, error patterns. Augmenting visual memory with question-aware attention maps yields notable gains on MMMU, but only marginal improvements on MathVista, because diagram-based tasks require more fine-grained visual understanding, e.g., smaller-scale vertex attention and higher spatial precision. More detailed analyses are provided in the Appendix.


\subsection{Memory Usage Analysis}

Figure~\ref{fig:memory_analysis} analyzes memory usage patterns across all benchmarks. Visual memory generation dominates the error collection, accounting for 59\%–93\% of stored cases in Figure~\ref{fig:memory_analysis}(a), demonstrating that visual perception remains the primary bottleneck in multimodal reasoning. Despite this generation asymmetry, both streams contribute comparably during retrieval, indicating effective memory reuse. Figure~\ref{fig:memory_analysis}(b) further confirms consistent dual-stream coordination across all three MLLMs, as reflected by the distribution of translucent retrieval points along the diagonal, indicating balanced contributions from both visual and logical streams. Moreover, our memory mechanism is not biased toward any specific model, as all three models exhibit similar patterns of memory utilization.


\subsection{Cross-Model Memory Transfer}

To evaluate the reusability and composability of the dual-stream memory framework, we conduct cross-model memory transfer experiments where each solver retrieves memories generated by other models. As shown in Table~\ref{tab:cross_model}, the 8B model benefits most from cross-model memories (+1.36 on {MMMU}, +1.33 on {MathVista}), surpassing its self-generated performance, indicating that memories distilled from stronger models encode higher-quality error patterns and generalization strategies. In contrast, larger models show comparable or slightly reduced performance, as their reasoning capabilities already yield near-optimal memory formation. These results highlight that dual-stream memory supports effective knowledge distillation from stronger to weaker models, enabling collaborative learning without explicit fine-tuning or ensembling.

\subsection{Cross-Benchmark Memory Generalization}
We assess memory transferability across task domains using {Qwen3-VL-8B-Instruct}. For each target benchmark in Table~\ref{tab:cross_benchmark}, we exclude its task-specific memory bank and instead retrieve from memories accumulated across \textit{all other benchmarks}. The results reveal substantial heterogeneity: {MathVision} and {RealWorldQA} benefit from cross-domain memories, as both require strong spatial reasoning. In contrast, tasks with large domain gaps, such as {MathVista} and {HallusionBench} (diagram-grounded vs. natural image reasoning), exhibit conflicts in memory utilization. Overall, the persistent gap between cross-domain and \textbf{ViLoMem} underscores that task-aligned memories are essential for optimal performance, validating our design choice to maintain distinct memory banks for different domains.


\subsection{{Memory Scalability}}
\label{sec:scalability}

{To stress-test long-term memory growth, we construct a progressive memory pool by sequentially accumulating memories from four math-domain benchmarks:
MathGlance~\citep{sun2025mathglance}, MathVista~\cite{lumathvista}, MathVision~\cite{wang2024measuring}, and MathVerse~\cite{zhang2024mathverse}. This sequence follows a visual-to-reasoning progression: we first build memory on diagram perceptual tasks (MathGlance) and then transition to visual grounded reasoning  tasks. The resulting pool contains ${\sim}$3k samples and 150k memory tokens. We then evaluate on the \textit{unseen} WeMath~\citep{qiao2025we}. This setup simulates a realistic long-horizon deployment where the memory pool grows incrementally across diverse tasks.}

\begin{table}[tb]
\centering
\caption{{Memory scalability on WeMath. Memory is progressively accumulated from four math-domain benchmarks (MathGlance$\rightarrow$MathVista$\rightarrow$MathVision$\rightarrow$MathVerse) and evaluated on the unseen WeMath. ``Direct'' denotes memory generated directly on WeMath itself.}}
\label{tab:scalability}
\vspace{-.6em}
\tablestyle{2pt}{1.}
\resizebox{.48\textwidth}{!}{
\begin{tabular}{lccccc|c}
\toprule
\#Tokens (samples) & 15k (0.1k) & 50k (0.5k) & 87k (1k) & 105k (2k) & 150k (3k) & Direct \\
\midrule
WeMath Acc. (\%) & 72.53 & 72.82 & 73.91 & 74.33 & \textbf{74.58} & 73.85 \\
\bottomrule
\end{tabular}}
\vspace{-2.2em}
\end{table}

{As shown in Table~\ref{tab:scalability}, accuracy on WeMath consistently improves as memory scales from 15k to 150k tokens (72.53$\rightarrow$74.58), demonstrating effective long-term memory scaling without catastrophic forgetting. Notably, cross-benchmark progressive memory (74.58) even outperforms memory generated directly on WeMath itself (73.85, ``Direct''), validating that abstract reasoning patterns accumulated across diverse math tasks transfer effectively to unseen problems.} 

\vspace{-.2em}
\section{Conclusion}
\vspace{-.2em}
\label{sec:con}

We introduce \textbf{ViLoMem}, a dual-stream memory framework that separately models visual distraction patterns and logical hallucination errors for multimodal large language models. Inspired by human semantic memory systems, \textbf{ViLoMem} coordinates visual and logical memory streams through specialized retrieval strategies and grow-and-refine update mechanisms. Comprehensive evaluations across six multimodal benchmarks demonstrate consistent improvements, with particularly pronounced gains on mathematical reasoning tasks where visual-logical coupling is most acute. Ablation studies confirm that both memory streams are complementary; joint operation enables synergistic error correction. Further analyses reveal heterogeneous cross-domain transfer behavior—task-aligned domains benefit from shared memory, whereas domain-mismatched tasks exhibit mild interference. Moreover, cross-model transfer experiments highlight that our memory can distill error patterns and reasoning strategies from stronger models to smaller ones, demonstrating its potential as a lightweight knowledge-sharing mechanism without explicit fine-tuning.  By enabling progressive error reduction without catastrophic forgetting, \textbf{ViLoMem} builds a foundation for continual learning in multimodal reasoning.

\section*{Acknowledgment}
This work was supported by National Natural Science Foundation of China (Grant No. 62425603).



\clearpage
\setcounter{page}{1}
\maketitlesupplementary

\section{Additional Results and Ablation Study}
\label{sec:additional_results}

\subsection{Integration with more models}
\begin{table*}[tb]
\centering
\caption{Additional evaluation results on GLM4.1v, InternVL3-38B, and Gemini2.5-flash across six multimodal reasoning benchmarks. Models with ``(step)'' and ``(+ \textbf{ViLoMem})'' are prompted by step-by-step reasoning. Results demonstrate consistent improvements from \textbf{ViLoMem} across diverse model architectures.}
\vspace{-.5em}
\label{tab:additional_models}
\resizebox{\textwidth}{!}{
\tablestyle{6pt}{1.1}
\begin{tabular}{l|cccccc}
\toprule
\textbf{Method} & \textbf{MMMU (dev)} & \textbf{MathVista (mini)} & \textbf{MathVision (mini)} & \textbf{HallusionBench} & \textbf{MMStar} & \textbf{RealWorldQA} \\
\midrule
GLM4.1v (baseline) & 69.14 & 72.57 & 56.88 & 73.08 & 72.90 & 73.33 \\
GLM4.1v (step) & 70.29 & 73.47 & 58.22 & 72.77 & 73.40 & 72.54 \\
\rowcolor{lightgold} GLM4.1v (+ \textbf{ViLoMem}) & \textbf{71.52} & \textbf{73.97} & \textbf{61.51} & \textbf{74.02} & \textbf{73.47} & \underline{72.68} \\
\midrule
InternVL3-38B (baseline) & 62.92 & 70.80 & 35.53 & 67.40 & 69.33 & 71.99 \\
InternVL3-38B (step) & 64.18 & 71.90 & 35.56 & 71.50 & 67.80 & 72.42 \\
\rowcolor{lightgold} InternVL3-38B (+ \textbf{ViLoMem}) & \textbf{65.97} & \textbf{73.80} & \textbf{36.84} & \textbf{72.34} & \textbf{69.73} & \textbf{73.20} \\
\midrule
Gemini2.5-flash (baseline) & 72.18& 81.10 & 53.21& 72.67 & 72.07& 76.99 \\
Gemini2.5-flash (step) & 71.90& 81.41& 53.94& 76.34& 72.40& 71.50 \\
\rowcolor{lightgold} Gemini2.5-flash (+ \textbf{ViLoMem}) & \textbf{72.86}& \textbf{83.40} & \textbf{58.22}& \textbf{78.33}& \textbf{73.20}& \underline{76.42} \\
\bottomrule
\end{tabular}
}
\vspace{-1.em}
\end{table*}

To verify the flexibility of \textbf{ViLoMem}, we extend our evaluation beyond the main experiments to recent reasoning-enhanced models, including GLM-4.1v~\cite{hong2025glm}, InternVL3-38B~\cite{zhu2025internvl3exploringadvancedtraining}, and Gemini 2.5~\cite{comanici2025gemini}. As shown in Table~\ref{tab:additional_models}, \textbf{ViLoMem} demonstrates robust adaptability across different architecture designs and inference regimes, consistently improving performance over both baseline and step-by-step configurations. This pattern echoes our observations in the main paper that visual perception remains a dominant bottleneck for multimodal reasoning~\citep{lumathvista,zhang2025primitive} and that decoupling visual distraction from logical hallucination yields complementary gains across tasks. Notably, models equipped with ``thinking'' or long-chain reasoning capabilities exhibit superior compatibility with the step-by-step format required for memory retrieval: their extended inference process allows for tighter integration of retrieved visual and logical guidelines into the reasoning chain, enabling them to correct potential errors before they propagate. These results suggest that \textbf{ViLoMem} is particularly well-suited to models with strong deliberative reasoning, while still offering consistent benefits to smaller or less capable solvers.

\begin{table*}[htb]
    \centering
    \caption{Comprehensive ablation study and comparison with existing memory methods across six multimodal reasoning benchmarks. We compare \textbf{ViLoMem} with attention mechanism variants and the Dynamic-Cheetsheet~\citep{suzgun2025dynamic} baseline adapted for multimodal tasks.}
    \label{tab:suppl_ablation}
    \resizebox{\textwidth}{!}{
    \tablestyle{6pt}{1.1}
    \begin{tabular}{l|cccccc}
    \toprule
    \textbf{Method} & \textbf{MMMU (dev)} & \textbf{MathVista (mini)} & \textbf{MathVision (mini)} & \textbf{HallusionBench} & \textbf{MMStar} & \textbf{RealWorldQA} \\
    \midrule
    \multicolumn{7}{l}{\textit{GPT-4.1}} \\
    \quad baseline & 74.00 & 70.40 & 46.12* & 58.50 & 69.80 & 73.72 \\
    \quad step & 74.16 & 74.27 & 47.47 & 74.44 & 70.43 & 72.03 \\
    \quad + dynamic-cheetsheet & 70.95 & 73.87 & 48.68 & 75.30 & 68.68 & 70.13 \\
    \rowcolor{lightgold} \quad + \textbf{ViLoMem} & 77.26 & \textbf{76.88} & \textbf{53.95} & 75.29 & \textbf{72.43} & \textbf{74.38} \\
    \rowcolor{moonveil} \quad + \textbf{ViLoMem} \& attention & \textbf{78.21} & 76.87 & 50.66 & \textbf{75.73} & 71.76 & \textbf{74.38} \\
    \midrule
    \multicolumn{7}{l}{\textit{Qwen3-VL-235B-A22B-Instruct}} \\
    \quad baseline & 78.70 & 84.90 & 61.28* & 63.20 & 78.40 &  79.30 \\
    \quad step & 75.97 & 83.66 & 62.17 & 74.58 & 76.16 & 78.66 \\
    \quad + dynamic-cheetsheet & 72.13 & 83.25 & 60.06 & 70.62 & 75.49 & 77.11 \\
    \rowcolor{lightgold} \quad + \textbf{ViLoMem} & \textbf{79.40} & \textbf{84.98} & \textbf{62.83} & 75.21 & 78.31 & 77.22 \\
    \rowcolor{moonveil} \quad + \textbf{ViLoMem} \& attention & 78.14 & 83.87 & 60.86 & \textbf{75.95} & \textbf{78.46} & \underline{77.88} \\
    \midrule
    \multicolumn{7}{l}{\textit{Qwen3-VL-8B-Instruct}} \\
    \quad baseline & 66.38* & 77.20 & 48.13* & 61.10 & 70.91 & 71.50 \\
    \quad step & 65.52 & 77.80 & 48.35 & 73.08 & 70.22 & 70.85 \\
    \quad + dynamic-cheetsheet & 63.39 & 74.92 & 46.81 & 68.39 & 69.12 & 69.98 \\
    \rowcolor{lightgold} \quad + \textbf{ViLoMem} & \textbf{69.90} & \textbf{77.87} & \textbf{49.34} & 73.19 & \textbf{72.13} & \textbf{73.59} \\
    \rowcolor{moonveil} \quad + \textbf{ViLoMem} \& attention & 67.52 & 77.07 & 48.72 & \textbf{74.87} & 72.67 & 73.46 \\
    \bottomrule
    \end{tabular}
    }
    \vspace{-1.em}
\end{table*}

\subsection{Attention Mechanism Ablation}
Table~\ref{tab:suppl_ablation} presents the ablation study of the attention mechanism. In general, the integration of attention maps yields consistent performance gains across hallucination and general reasoning benchmarks (e.g., HallusionBench, MMStar), corroborating the critical importance of visual memory in refining perceptual grounding. However, we observe a performance plateau or marginal decline on mathematics-centric datasets (MathVista and MathVision). We attribute this limitation to two primary factors: (1) \textbf{Visualization Precision}: Current attention visualization methods struggle to faithfully preserve fine-grained geometric structures and chart details, which are essential for mathematical reasoning. (2) \textbf{Contextual Interpretation}: While serving as an auxiliary image to enhance visual context, the attention map imposes higher demands on the model's intrinsic capability to interpret heatmap overlays. The benefit of this enriched context is contingent on the model's ability to align these explicit visual cues with the raw image features without information loss.

\subsection{Additional Case Study}
\begin{figure*}[!htb]
\centering
\includegraphics[width=0.95\textwidth]{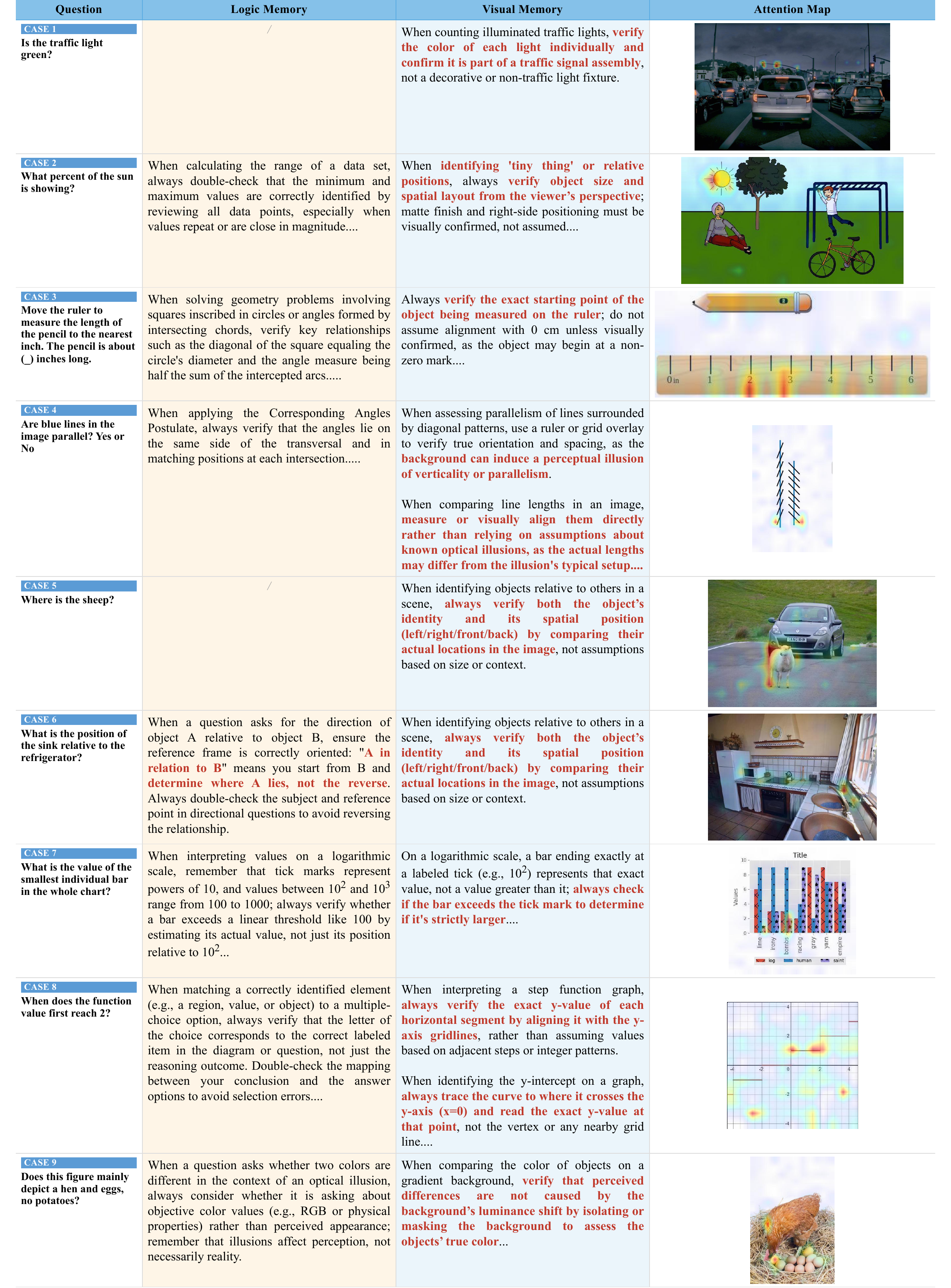}
\caption{Showcase of representative cases demonstrating \textbf{ViLoMem}'s memory generation and retrieval process across different types of multimodal reasoning tasks.}
\label{fig:appendix_cases}
\vspace{-1em}
\end{figure*}

Figure~\ref{fig:appendix_cases} summarizes representative qualitative cases. For many vision-intensive questions (e.g., traffic-light color, visible portion of the sun, object localization, and optical-illusion setups), logical memory is either not retrieved or fails to offer useful guidance, while visual memory provides concrete viewing strategies such as checking the actual illuminated region, reading tiny objects and relative positions from the viewer's frame, or isolating targets from distracting backgrounds. In these cases, attention maps concentrate on the queried regions (e.g., the active light, visible solar arc, or relevant segments), so that the retrieved visual guidelines directly steer the solver toward task-relevant evidence. 

For geometry and chart-reading tasks, visual and logical memories are complementary: logical memory provides reusable rules for measurement and graph interpretation, while visual memory focuses on concrete inspection behaviors such as aligning with gridlines, following step edges, or checking true line orientation under strong illusions. Together, these cases highlight a clear division of labor: visual memory governs ``where to look'' and mitigates systematic perceptual traps, whereas logical memory refines ``how to reason'' once the correct visual evidence has been attended.

\subsection{Comparison with Existing Memory Methods}
We benchmark \textbf{ViLoMem} against state-of-the-art memory mechanisms~\citep{suzgun2025dynamic,zhang2025agentic}. While the original Dynamic-Cheetsheet~\citep{suzgun2025dynamic} employs cumulative memory, its unbounded context growth is infeasible for our large-scale setting (approx.\ 1,000 cases per benchmark), so we adopt the retrieval-based configuration from the open-source Dynamic-Cheetsheet codebase, which follows the similar methodology as ACE~\citep{zhang2025agentic}. For a fair multimodal comparison, we replicate the official prompt structure and use the same MLLM for both memory generation and inference. In this setup, the retrieval module relies purely on text similarity without image-aware matching.

Experimental results in Table~\ref{tab:suppl_ablation} show that this direct adaptation of logical memory methods is suboptimal in multimodal settings and can even underperform the baseline, especially for smaller models. In practice, such text-only retrieval often surfaces visually dissimilar examples with similar questions, resurfacing prior misperceptions as salient ``hints'' that misdirect attention away from the correct regions of the current problem. Qualitative inspection further reveals that Dynamic-Cheetsheet and ACE are tailored to code- or logic-centric schemas: even when driven by an MLLM, they mainly produce fine-grained corrections of specific visual details (digits, colors, marks) rather than robust guidance on how to inspect diagrams. These detail-level cues lack stable visual grounding and easily conflict with the actual image, inducing additional hallucinations that smaller models are particularly vulnerable to. This contrast highlights the need for \textbf{ViLoMem}'s decoupled visual stream and question-aware retrieval, which explicitly organize and retrieve perception-oriented error patterns instead of repurposing logic-only memories.

\noindent\textbf{Retrieval Efficiency at Scale.}
We further analyze how the two-stage retrieval pipeline scales with growing memory. Table~\ref{tab:retrieval_efficiency} compares accuracy and per-case latency (ms) on WeMath for: no memory (S0), visual-only retrieval (S1), text-only retrieval (S2), and two-stage retrieval (S$_{\oplus}$ = S1 + S2). As memory grows from 87k to 150k tokens, S2 latency increases from 22.8ms to 115ms due to full-corpus text matching, while S$_{\oplus}$ latency only rises from 20.4ms to 61.6ms, a 22.3$\times$ reduction compared to S2 alone. Due to S1 first narrows the candidate set via efficient image embedding search (1024-dim), allowing S2 to rerank only the top candidates. Crucially, S$_{\oplus}$ also achieves the highest accuracy (74.58\%), as the two stages provide complementary filtering: S1 captures visual similarity while S2 refines by semantic relevance.

\begin{table}[tb]
\centering
\caption{Two-stage retrieval analysis on WeMath. S0: no memory (baseline); S1: visual retrieval; S2: text retrieval; S$_{\oplus}$: S1 + S2 (two-stage). Latency in ms per case.}
\label{tab:retrieval_efficiency}
\vspace{-.5em}
\tablestyle{4pt}{1.}
\resizebox{.48\textwidth}{!}{
\begin{tabular}{lcccc}
\toprule
Stage / Acc.\ \% {\scriptsize(Latency ms)} & S0 & S1 & S2 & S$_{\oplus}$ \\
\midrule
87k tokens
& 72.07
& 72.94 {\scriptsize(19.1)}
& {73.22} {\scriptsize(22.8)}
& {73.91} {\scriptsize(20.4)} \\
150k tokens
& 72.07
& 73.72 {\scriptsize(60.7)}
& {73.41} {\scriptsize(115)}
& {\textbf{74.58}} {\scriptsize(61.6)} \\
\bottomrule
\end{tabular}}
\vspace{-1em}
\end{table}

We further compare the computational efficiency of \textbf{ViLoMem} against Dynamic-Cheetsheet (DC) on MathVista in Table~\ref{tab:efficiency_comparison}. \textbf{ViLoMem} achieves higher accuracy (+3.01) while requiring significantly lower retrieval latency ($-$63.1\%) and storage cost ($-$66.8\%). This efficiency stems from our selective memory update strategy: \textbf{ViLoMem} only generates memory entries for incorrect answers, whereas DC updates memory for every case, leading to a bloated memory pool (221K vs.\ 73K tokens) with proportionally higher retrieval overhead.

\begin{table}[tb]
\centering
\caption{Efficiency comparison between ViLoMem and Dynamic-Cheetsheet (DC) on MathVista. ViLoMem achieves superior accuracy with substantially lower retrieval cost and storage.}
\label{tab:efficiency_comparison}
\vspace{-.5em}
\tablestyle{2pt}{1.1}
\resizebox{.48\textwidth}{!}{
\begin{tabular}{lcccc}
\toprule
Method & Acc. (\%) & Retrieval (ms) & Storage (MB) & \#Memory (tokens) \\
\midrule
Baseline & 70.40 & 0 & 0 & 0 \\
DC & 73.87 & 325 & 18.44 & 221K \\
\textbf{ViLoMem} & \textbf{76.88} & \textbf{120} & \textbf{6.12} & \textbf{73K} \\
\bottomrule
\end{tabular}}
\vspace{-1em}
\end{table}

\subsection{Failure Case Analysis}
\label{sec:failure_analysis}

To understand the limitations of \textbf{ViLoMem}, we analyze cases on MMMU (1,041 samples, GPT-4.1 as solver) where the baseline answers correctly but \textbf{ViLoMem} fails. Overall, \textbf{ViLoMem} achieves a net gain of +8.86\% over baseline; however, we identify 66 regression cases where memory retrieval hurts performance. Among these, 33.3\% (22 cases) are caused by \textit{generic visual memory}: the retrieved visual cues are only weakly relevant to the specific image and question, lacking image-specific and question-specific adaptation, which distracts reasoning despite being knowledge-correct. The remaining 66.7\% (44 cases) are caused by \textit{empty retrieval}: no memory is retrieved at all (0 matched entries), meaning the model receives no memory-augmented guidance and instead relies on a step-by-step prompt that may differ from the baseline's default prompting. Notably, no failures occur when both visual and logical memory are successfully retrieved, suggesting that the dual-stream retrieval mechanism is reliable when sufficient memory coverage exists.

\subsection{Visual Memory Characterization}
\label{sec:visual_memory_characterization}

As shown in Table~\ref{tab:ablation}, \textbf{ViLoMem} achieves 77.26\% on MMMU with GPT-4.1, with both visual and logical  memory contributing to the overall gain. To understand \textit{which types of tasks} benefit most from each memory stream, we further analyze the 835 subject-matched samples across MMMU's six academic disciplines. Table~\ref{tab:visual_characterization} reports the improvement from the full system over baseline ($\Delta$(B$\rightarrow$VLM)), and the independent contribution of each memory stream: $\Delta$(vis) = ViLoMem $-$ w/o visual, and $\Delta$(log) = ViLoMem $-$ w/o logic.

\begin{table}[tb]
\centering
\caption{Per-discipline memory contribution analysis on MMMU (GPT-4.1, 835 subject-matched samples). $\Delta$(vis) and $\Delta$(log) measure the independent contribution of visual and logical memory, respectively. Bold indicates the dominant memory stream.}
\label{tab:visual_characterization}
\vspace{-.5em}
\tablestyle{3pt}{1.1}
\resizebox{.48\textwidth}{!}{
\begin{tabular}{l|c|c|cc}
\toprule
\textbf{Discipline} & \textbf{N} & $\boldsymbol{\Delta}$\textbf{(B$\rightarrow$VLM)} & $\boldsymbol{\Delta}$\textbf{(vis)} & $\boldsymbol{\Delta}$\textbf{(log)} \\
\midrule
Tech \& Engineering & 205 & +12.7 & \textbf{+9.8} & +5.4 \\
Health \& Medicine & 99 & +10.1 & +1.0 & \textbf{+8.1} \\
Humanities \& Social Sci. & 137 & +9.5 & +0.7 & +0.0 \\
Science & 136 & +8.1 & +3.7 & \textbf{+4.4} \\
Art \& Design & 93 & +3.2 & +1.1 & \textbf{+4.3} \\
Business & 165 & +0.0 & $-$3.6 & $-$4.2 \\
\bottomrule
\end{tabular}}
\vspace{-1em}
\end{table}

The results reveal clear discipline-dependent memory utilization patterns:

\noindent\textbf{Visual memory dominates in Tech \& Engineering} ($\Delta$(vis)=+9.8 vs.\ $\Delta$(log)=+5.4). At the subject level, Energy \& Power (+20.0), Math (+19.4), Agriculture (+14.3), and Mechanical Engineering (+12.5) benefit most from visual memory. These subjects feature specialized diagrams (circuit schematics, engineering drawings, mathematical plots, microscopic images) that require recognizing domain-specific visual patterns beyond the model's general vision capability.

\noindent\textbf{Logical memory dominates in Health \& Medicine} ($\Delta$(log)=+8.1 vs.\ $\Delta$(vis)=+1.0). Diagnostics \& Laboratory Medicine shows the largest overall gain (+24.2 from baseline), but the improvement is primarily driven by logical memory (+9.1), indicating that medical reasoning chains---rather than visual perception---are the main bottleneck.

\noindent\textbf{Both streams are ineffective for Business} ($\Delta$(vis)=$-$3.6, $\Delta$(log)=$-$4.2). The model's baseline visual encoder already handles standard business charts and tables well, and the retrieved memory introduces noise that harms performance. This highlights that memory augmentation is most valuable when the task exceeds the model's inherent capabilities.

\subsection{Results on Visual Perception Benchmarks}
\label{sec:perception_benchmarks}

 On the math perception benchmark MathGlance~\citep{sun2025mathglance}, \textbf{ViLoMem} generates 166 visual and 16 logic memory entities, achieving +1.87\% average accuracy improvement over the no-memory baseline on plane/solid geometry and graph tasks. Note that purely perceptual benchmarks such as OCRBench and DocVQA are not directly applicable to our framework, as they use scoring-based evaluation (character matching) rather than binary correctness judgments, which prevents the error-driven memory generation process.

\section{Additional Experimental Details}
\label{sec:additional_details}

This section provides additional implementation details that complement the experimental setup.

\noindent\textbf{Model Deployment.}
For open-source models, we deploy most checkpoints using \texttt{vLLM} for efficient batched inference.
Due to its scale, \textit{Qwen3-VL-235B-A22B-Instruct} is accessed via its official API instead of local deployment, and all proprietary models (e.g., GPT-4.1, Gemini 2.5 flash) are evaluated through their corresponding APIs.
For API-based evaluations, certain images or prompts may be flagged as unsafe by the provider's safety filters and thus rejected, which introduces a small amount of noise into the reported scores.

\noindent\textbf{Decoding Hyperparameters.}
Unless otherwise specified, we use a temperature of $0.7$ and a maximum generation length of $8{,}192$ tokens for all models.
Within our memory pipeline, the maximum generation length is set to $1{,}024$ tokens for problem analysis and $2{,}048$ tokens for memory generation to balance expressiveness and efficiency.
Baseline evaluations directly feed benchmark questions to the models without additional prompts, whereas the \textit{Step} configuration prepends a simple step-by-step system prompt; the full template is shown in Figure~\ref{fig:prompt_step}.

\noindent\textbf{Attention Map Generation.}
Attention maps are generated following the training-free small-detail perception framework of Zhang et al.~\citep{zhangmllms}, instantiated with \textit{Qwen2.5-VL-3B} as the backbone model.
This setup produces token-level saliency over input images, which we overlay as heatmaps to visualize and interpret visual memory retrieval.

\noindent\textbf{Evaluation Protocol.}
We adopt VLMEvalKit~\citep{duan2024vlmevalkit} as the primary evaluation framework.
When automatic matching fails or produces ambiguous results (e.g., due to formatting variations), we further apply \textit{Math-Verify} and an LLM-as-a-judge protocol to reduce sensitivity to output formatting.
The judge model is \textit{Qwen3-8B-Instruct}, which assesses whether a model's response is semantically correct with respect to the reference answer.

\begin{figure*}[htb]
\centering
\begin{tcolorbox}[colback=gray!5,colframe=gray!40!black,title=\textbf{Prompt: Step-by-Step Reasoning}]
\small
\textbf{Objective:} Solve the given problem using a step-by-step process.

\vspace{0.5em}
\textbf{Expected Output Structure:}
\begin{verbatim}
Step 1:
Step 2:
...
Step n: Final Answer: \boxed{answer}

Question:
\end{verbatim}
\end{tcolorbox}
\caption{The step-by-step reasoning system prompt used in the \textit{Step} configuration.}
\label{fig:prompt_step}
\end{figure*}

\begin{figure*}[!htb]
\centering
\begin{tcolorbox}[colback=gray!5,colframe=gray!40!black,title=\textbf{Prompt: Problem Analysis}]
\small
\textbf{Objective:}

Analyze the following problem to identify its subject area and the key concepts, principles, formulas, or laws required for its solution. This analysis will be used to retrieve relevant guiding principles from a knowledge base.

\textbf{Instructions:}
\begin{itemize}
    \item Do not solve the problem.
    \item First, identify the primary subject (e.g., Physics, Chemistry, Biology, Mathematics).
    \item Then, list the core concepts or principles involved (e.g., Newton's Second Law, Conservation of Energy, Stoichiometry, Pythagorean theorem).
    \item Keep the analysis concise and focused.
\end{itemize}

\textbf{Problem:}

\{question\}

\textbf{Output Format:}

Subject: $<$The primary subject$>$

Key Concepts: $<$A brief list of key concepts$>$
\end{tcolorbox}
\caption{The prompt template for analyzing the problem to identify its subject and key concepts.}
\label{fig:prompt_analyze}
\end{figure*}

\begin{figure*}[t]
\centering
\begin{tcolorbox}[colback=gray!5,colframe=gray!40!black,title=\textbf{Prompt: Logical Memory Generation}]
\small
\textbf{Objective:}

Analyze the provided incorrect reasoning process for a scientific or mathematical problem. Your goal is to classify the error and, if it is a logical error, generate a high-quality, actionable guideline (a ``memory'') to prevent similar mistakes in the future.

\textbf{Context:}
\begin{itemize}
    \item Problem: \{question\}
    \item Incorrect Reasoning Steps: \{reasoning\_steps\}
    \item Correct Answer (for reference): \{gold\_answer\}
\end{itemize}

\textbf{Instructions:}

\begin{enumerate}
    \item \textbf{Analyze the Mistake:} Carefully review the \texttt{Incorrect Reasoning Steps} against the \texttt{Problem} and \texttt{Correct Answer} to pinpoint the primary mistake.
    \item \textbf{Categorize the Error:} Classify the error into one of two types:
    \begin{itemize}
        \item \texttt{Logical}: Any error in the reasoning process itself. This includes calculation mistakes, misapplication of a formula or theorem, logical fallacies, or conceptual misunderstandings. These errors can be identified from the text of the reasoning alone.
        \item \texttt{Non-Logical}: An error that stems purely from misinterpreting the visual information in an image. This kind of error can \textbf{only} be confirmed by looking at the image (e.g., misidentifying a shape, reading a value from a graph incorrectly).
    \end{itemize}
    \item \textbf{Generate the Guideline (Memory):}
    \begin{itemize}
        \item \textbf{Only if the \texttt{error\_type} is \texttt{Logical}}, you must generate a guideline.
        \item If the \texttt{error\_type} is \texttt{Non-Logical}, the guideline must be left empty.
    \end{itemize}

    \textbf{Guideline Quality Requirements:}
    \begin{itemize}
        \item \textbf{Be Specific and Concrete:} The guideline must target the specific principle, theorem, formula, or reasoning pattern that was misused. Name the concept directly.
        \item \textbf{Be Actionable:} Frame it as a clear instruction, a warning, or a rule of thumb (e.g., ``Always check...'', ``Remember to differentiate between...'', ``When X occurs, apply Y...'').
        \item \textbf{Be Generalizable:} The advice should be abstracted from the specific numbers and context of this single problem so it can apply to a whole class of similar problems.
        \item \textbf{Keep it Concise:} The guideline should be one to two sentences long.
    \end{itemize}

    \textbf{Guideline Examples (Good, Specific Examples):}
    \begin{itemize}
        \item \textbf{(Physics):} ``When applying the conservation of energy to rolling objects, always include both translational and rotational kinetic energy in the equation.''
        \item \textbf{(Math):} ``In geometry problems involving tangents to a circle, remember that the radius to the point of tangency is perpendicular to the tangent line.''
        \item \textbf{(Chemistry):} ``For stoichiometry calculations, always ensure the chemical equation is correctly balanced before determining the molar ratios.''
    \end{itemize}
\end{enumerate}

\textbf{Output Format (use this exact structure):}

error\_type: $<$``Logical'' or ``Non-Logical''$>$

analysis\_summary: $<$A brief, one-sentence summary of what went wrong.$>$

guideline: $<$Your 1-2 sentence guideline if the error is ``Logical'', otherwise leave this field empty.$>$
\end{tcolorbox}
\caption{The prompt template for generating logical memories.}
\label{fig:prompt_logical}
\end{figure*}

\begin{figure*}[htb]
\centering
\begin{tcolorbox}[colback=gray!5,colframe=gray!40!black,title=\textbf{Prompt: Visual Memory Generation}]
\small
\textbf{Objective:}

You are an expert in visual reasoning and error analysis. Your task is to first describe the provided image objectively, then analyze an incorrect reasoning process to determine if the error stems from misinterpreting that image. If a visual error is found, you must generate a concise, actionable guideline (a ``visual memory'') to prevent this mistake in the future.

\textbf{Context:}
\begin{itemize}
    \item Problem: \{question\}
    \item Incorrect Reasoning Steps: \{reasoning\_steps\}
    \item Correct Answer (for reference): \{gold\_answer\}
\end{itemize}

\textbf{Attached Image:} $<$image$>$

\textbf{Thinking Process and Final Output:}

Your response must follow a strict two-stage process. The first stage is your internal ``thought process'' which you will write out. The second stage is the final JSON output.

\textbf{Stage 1: Internal Thought Process (Write this out first)}
\begin{enumerate}
    \item \textbf{Describe the Image:} Begin by providing an objective, detailed description of the attached image. List all key elements, labels, values, geometric shapes, and their relationships. This description will serve as the ``ground truth'' for your analysis.
    \item \textbf{Analyze for Discrepancies:} Compare your image description and the image itself against the text in \texttt{Incorrect Reasoning Steps}. Identify any contradictions, misinterpretations, or omissions.
\end{enumerate}

\textbf{Stage 2: Final JSON Output (Provide ONLY this JSON block as the final answer)}

After completing your thought process, generate a JSON object based on your analysis. The JSON should adhere to the following structure and guidelines.

\textbf{Guidelines for \texttt{guideline} Generation:}
\begin{itemize}
    \item The guideline MUST be about how to correctly interpret a specific visual pattern or element.
    \item It must be a rule that can be applied to other, similar-looking problems.
    \item It should be concise (one to two sentences).
\end{itemize}

\textbf{Guideline Examples (Good, Specific Visual Memories):}
\begin{itemize}
    \item \textbf{(Physics/Diagrams):} ``In a free-body diagram, always verify that all forces, including friction and normal force, are accounted for before applying Newton's laws.''
    \item \textbf{(Geometry):} ``When an angle appears to be a right angle in a diagram, do not assume it is 90 degrees unless it is explicitly marked with a square symbol.''
    \item \textbf{(Chemistry/Molecules):} ``For complex organic molecules, double-check the placement of double bonds and functional groups as they dictate the molecule's reactivity.''
    \item \textbf{(Biology/Graphs):} ``When reading a bar chart, pay close attention to the Y-axis scale and units to avoid misinterpreting the magnitude of the results.''
\end{itemize}

\textbf{Avoid these types of guidelines (Bad, Non-Visual or Too Vague):}
\begin{itemize}
    \item ``The model made a calculation error.'' (This is a logical error, not visual)
    \item ``You need to look at the image more carefully.'' (Not actionable)
    \item ``The reasoning about the physics was wrong.'' (Too general)
\end{itemize}

\textbf{Final Output Format (use this exact JSON structure):}
\begin{verbatim}
{
    "is_visual_error": true/false,
    "analysis_summary": "A brief, one-sentence summary of the visual 
                        misinterpretation.",
    "guideline": "Your 1-2 sentence visual guideline. Provide this 
                 only if is_visual_error is true, otherwise it 
                 should be null."
}
\end{verbatim}
\end{tcolorbox}
\caption{The prompt template for generating visual memories.}
\label{fig:prompt_visual}
\end{figure*}

\begin{figure*}[htb]
\centering
\begin{tcolorbox}[colback=gray!5,colframe=gray!40!black,title=\textbf{Prompt: LLM-as-a-Judge Verification}]
\small
\textbf{Objective:}

You are an expert answer verification judge. Your task is to determine whether a model prediction matches the gold answer.

\vspace{0.5em}
\textbf{Core Principle (Critical Rule):}
\begin{itemize}
    \item All decisions are based \emph{only} on the gold answer; ignore the quality of the reasoning.
    \item If the extracted final answer from the prediction exactly matches the gold answer, set \texttt{verified=true}; otherwise, set \texttt{verified=false}.
    \item Do not consider whether the prediction's reasoning is correct or sensible.
    \item Do not give partial credit for ``close'' answers (e.g., $2 \neq 9$, C $\neq$ A).
\end{itemize}

\textbf{Verification Steps:}
\begin{enumerate}
    \item \textbf{Identify the gold answer format.}
    Determine whether the gold answer is:
    \begin{itemize}
        \item a single letter (\texttt{A/B/C/D/E}) for multiple-choice questions (compare letters only);
        \item a number for numerical questions (compare numeric values, ignoring formatting such as 7 vs.\ 7.0);
        \item a text span for open-ended questions (compare semantic meaning, allowing minor wording differences).
    \end{itemize}
    \item \textbf{Extract the final answer from the prediction.}
    \begin{itemize}
        \item For multiple-choice questions, locate the final chosen letter (often after ``Final Answer:'' or ``Answer:'') and compare it with the gold letter.
        \item For numerical questions, locate the final numeric value, ignoring units and extra text, then compare it to the gold number.
        \item For text answers, extract the final answer phrase and compare its semantic meaning with the gold text (e.g., ``Yes, the baby is crawling to the right.'' matches ``Yes'').
    \end{itemize}
    \item \textbf{Apply the strict matching rule.}
    \begin{itemize}
        \item Only compare the final extracted answers.
        \item Do not use external knowledge to judge whether an answer is reasonable.
        \item If the extracted answer and the gold answer match under the appropriate format, output \texttt{verified=true}; otherwise, output \texttt{verified=false}.
    \end{itemize}
\end{enumerate}

\textbf{Input Fields:}
\begin{itemize}
    \item Question: \{question\}
    \item Gold Answer: \{gold\_answer\}
    \item Choices (optional, for multiple choice): \{choices\_text\}
    \item Prediction: \{prediction\}
\end{itemize}

\textbf{Output Format (JSON, exact structure):}
\begin{verbatim}
{
  "reasoning": "Step 1: Extract answer from prediction: [extracted_value]. 
Step 2: Compare with gold: [gold_value]. 
Step 3: Match result: [yes/no].",
  "verified": true or false
}
\end{verbatim}
\end{tcolorbox}
\caption{The LLM-as-a-judge prompt template used to verify whether a model prediction matches the gold answer, independent of reasoning quality.}
\label{fig:prompt_llm_judge}
\end{figure*}

\section{Prompt Templates}
\label{sec:system_prompts}

We provide the full prompt templates used in our framework, including the step-by-step reasoning prompt used in the \textit{Step} configuration (Figure~\ref{fig:prompt_step}), the Problem Analysis Prompt (Figure~\ref{fig:prompt_analyze}), the Logical Memory Generation Prompt (Figure~\ref{fig:prompt_logical}), and the Visual Memory Generation Prompt (Figure~\ref{fig:prompt_visual}), together with the LLM-as-a-judge verification prompt (Figure~\ref{fig:prompt_llm_judge}).

\end{document}